\documentclass[12pt]{article}
\usepackage{graphicx} 
\usepackage{amsmath}
\usepackage{amssymb}
\usepackage{amsthm}
\usepackage{mathtools}
\usepackage{subcaption}
\usepackage{tikz}
\usepackage{csquotes}
\usepackage{hyperref}
\usepackage{natbib}

\newcommand{\sgn}{\text{sgn}}

\theoremstyle{plain}
\newtheorem{theorem}{Theorem}[section]
\newtheorem{proposition}[theorem]{Proposition}

\theoremstyle{definition}
\newtheorem{definition}[theorem]{Definition}

\theoremstyle{remark}

\title{Peirce in the Machine:  How Mixture of Experts Models Perform Hypothesis Construction}
\author{Bruce Rushing}
\date{June 2024}

\begin{document}

\maketitle

\begin{abstract}
    Mixture of experts is a prediction aggregation method in machine learning that aggregates the predictions of specialized experts.  This method often outperforms Bayesian methods despite the Bayesian having stronger inductive guarantees.  We argue that this is due to the greater functional capacity of mixture of experts.  We prove that in a limiting case of mixture of experts will have greater capacity than equivalent Bayesian methods, which we vouchsafe through experiments on non-limiting cases.  Finally, we conclude that mixture of experts is a type of abductive reasoning in the Peircian sense of hypothesis construction.
\end{abstract}

\noindent \textbf{Word Count:  9100}

\section{Introduction}\label{sec:intro}

Machine learning has developed many powerful techniques for solving prediction problems over the past few decades.  Among those are prediction aggregation methods that seek to combine multiple machine learning models to make better predictions than can be achieved by individual predictors.  This has strong theoretical foundations from the theory of meta induction \citep{cesa2006prediction, schurz2019hume} as well as empirical support for the various methods \citep{dietterich2000ensemble, rokach2010ensemble, masoudnia2014mixture}.  Two important strategies for prediction aggregation are Bayesian Model Averages (BMA) and Mixtures of Experts (MOE).  The former is a Bayesian method to make predictions by averaging predictions across a posterior of possible models learned from the data; the latter involves learning specialized experts and then making predictions based on which experts would make the best prediction based on observed features.  BMA has been a widely preferred strategy because of strong theoretical guarantees about its inductive superiority.  But surprisingly, MOE often performs better in practice, which has led to its adoption as a crucial element in many cutting-edge machine learning models like large language models.  We aim to address this startling fact in this paper.

We argue that MOEs outperform BMAs because MOEs create more expressive hypotheses or models than BMAs.  It can be shown that in at least one exact sense of capacity, a MOE can have greater capacity than a BMA composed of similar models to that MOE.  We then validate this theoretical result with experiments that show as the capacity required to predict a dataset grows, MOEs maintain good performance while BMAs fall off based on the capacity of their members.  Both of these results suggest the superiority of MOEs over BMAs, despite the latter's inductive guarantees, stems from the restriction of BMAs to a poorer hypothesis class than MOEs.

The philosophical upshot of this finding is that MOEs can be thought of as doing abduction, in the sense of Peircian sense of hypothesis construction, along with induction.  In comparison, BMAs and other similar schemes are fundamentally inductive alone.  We argue that MOEs demonstrate a divide-and-conquer strategy for hypothesis formation, similar to a suggestion given by Peirce.  This means that MOE addresses a problem that Bayesian methods fail to address, which is where the hypothesis space comes from.

Here is how our argument proceeds.  First, we discuss MOEs as applied to supervised learning.  Second, we review BMAs, reasons for their optimality, and empirical evidence pushing against that optimality.  We discuss possible explanations for this difference but find them to be wanting.  Third, we propose another explanation centered on the capacity of MOEs compared to BMAs.  We show that with capacity made exact, we can prove that the capacity of MOEs will exceed the capacity of BMAs.  Fourth, we build on this theoretical result by demonstrating experiments that show MOE performance maintains itself on datasets that require a certain amount of capacity to learn well while BMAs drop off.  Fifth, we discuss the philosophical implications of this explanation, by highlighting that MOEs are essentially a form of abduction in the sense of hypothesis construction.  
\section{Mixtures of Experts}\label{sec:MOE}

In supervised learning, the goal of the machine learning algorithm is to predict some target $y$ from some vector of features $\mathbf{x}$, and this goal is realized by training the model on some dataset of $n$ target and feature pairs $\mathcal{D} = \{(\mathbf{x}^{(i)}, y^{(i)})\}_{i=1,\dots,n}$, where $\mathcal{X}$ and $\mathcal{Y}$ are the training features and targets respectively.  A model can be thought of as learning a probability function $f(y | \mathbf{x}, \theta)$ where $\theta$ are the model parameters, from which predictions can be sampled or deterministically picked as the target with the highest probability, i.e. the prediction $\hat{y} = \underset{y}{\arg \max} f(y | \mathbf{x}, \theta)$.  Training is accomplished by minimizing some loss or cost function $L(\theta)$ through a training algorithm to find the best set of model parameters $\theta$ for making predictions.  Models are then evaluated by how well they do on some hold-out test set $\mathcal{T} = \{(\mathbf{x}^{(i)}, y^{(i)})\}_{i=1,\dots,m}$ not seen during training.  The best models have the lowest loss on this test set.

For example, suppose the problem is to predict blood pressure from body mass index (BMI).  We treat our target $y$ as the continuous random variable blood pressure and our features $\mathbf{x} = [1, x_{1}]^{\intercal}$ as the continuous random variable BMI and identity feature.  A linear regression is then a normally distributed probability function $f_{lr}(y | \mathbf{x}, \theta)$ with known variance $\sigma^{2}$ and parameters $\theta = [\theta_{0}, \theta_{1}]^{\intercal}$ that characterize the mean of our distribution:

\begin{equation}\label{eq:linear-regression}
    f_{lr}(y | \mathbf{x}, \theta) = \mathcal{N}(\mathbf{x}^{\intercal}\theta, \sigma^{2})
\end{equation}

\noindent where $\mathbf{x}^{\intercal}$ is the transpose of our feature vector.  Assuming independence of our model parameters, we then fit our model by maximizing the likelihood $\hat{\theta} = \underset{\theta}{\arg \max}f_{lr}(\mathcal{Y} | \mathcal{X}, \theta)$, which corresponds to minimizing the negative log-likelihood loss function $L_{NLL}$:

\begin{equation}
    L_{NLL}(\theta) = - \overset{n}{\underset{i=1}{\sum}} \log f_{lr}(y^{(i)} | \mathbf{x}^{(i)}, \theta)
\end{equation}

\noindent Due to our model being normally distributed and ignoring constants that do not depend on $\theta$, the negative log-likelihood equates to minimizing the  squared error:

\begin{equation}
    L_{NLL}(\theta) = \frac{1}{2\sigma^{2}} \overset{n}{\underset{i=1}{\sum}} (y^{(i)} - \mathbf{x}^{(i)\intercal}\theta)^{2}
\end{equation}

\noindent We could then train our linear regression on the dataset $\mathcal{D}$ via gradient descent to find the parameters that minimize this loss function.  To evaluate how well our regression does, we evaluate its negative log-likelihood on a hold-out set of blood pressure and BMI pairs not seen during training.  This is done because we want our model to inductively generalize from its training; we don't want the model to have merely memorized the blood pressure and BMI pairs found in training.  The best model is one that successfully minimizes the training loss while having the lowest test set loss.

Aggregating predictions from many machine learning models is a common technique for improving predictive performance in supervised learning.  There are two primary families of aggregation:  ensemble methods and MOE methods.  

In ensemble methods, we train individual models independently across the feature set $\mathcal{X}$ and then use some aggregation scheme across all models to produce a prediction.  Examples of aggregation schemes include averaging and voting.  In the averaging case, the prediction of an ensemble machine learning model $g(\mathbf{x})$ is an average of $n$ models with probabilities $f(y | \mathbf{x}, \theta_{i})$ parameterized by $\theta_{i}$ and weighted by weights $w_{i}$ for $i = 1, \dots, n$:

\begin{equation}
    g(\mathbf{x}) = \underset{i=1}{\overset{n}{\sum}} w_{i}f(y | \mathbf{x}, \theta_{i})
\end{equation}

\noindent The core intuition behind ensemble techniques is that they leverage the wisdom of the crowds:  instead of relying upon one highly specialized model for prediction, many models are consulted and then factored into the prediction.  Betting markets in sports gambling can be thought of as an ensemble technique leveraging human gamblers as individual predictors, where the going price in the market corresponds to an average of all wagers.  Examples of ensemble techniques include bagging, where the random predictors are trained and weighted on random subsets of data, and boosting, where predictors are increasingly specialized on the errors of other predictors.

With a MOE \citep{jacobs1991adaptive, jordan1994hierarchical, yuksel2012twenty, masoudnia2014mixture}, models called experts specialize on elements of a partition of the feature set $\mathcal{X}$, and predictions on a new sample from the mixture consist of routing the sample to the relevant experts based on where in that partition the sample falls.  In the most general case where we route the sample to all experts, we employ weights for the experts that are a function of the sample features $\mathbf{x}$.  The MOE model $h(\mathbf{x}; n)$ with $n$ experts then is a sum over experts with probability functions $f_{i}(y | \mathbf{x}, \theta_{i})$ parameterized by $\theta_{i}$ and a gating function $G_{i}: \mathcal{X} \to [0,1]$ for $i = 1, \dots, n$:

\begin{equation}
    h(\mathbf{x}; n) = \underset{i=1}{\overset{n}{\sum}}G_{i}(\mathbf{x})f_{i}(y | \mathbf{x}, \theta_{i})
\end{equation}

\noindent The gating functions $G_{i}(\mathbf{x})$ are typically probabilities, such as the softmax over a linear model.  We say that experts are homogenous when expert probability functions $f_{i}$ have the same form but are characterized by different parameters $\theta_{i}$ and inhomogeneous otherwise.  For example, a MOE with only linear regressions as experts has homogenous experts.  Importantly, the gating function when it considers more than one expert in making a prediction, i.e. when more than one expert has positive probability for a given sample, is said to softly partition the feature set between experts (experts provide predictions for more than one partition cell).  The limiting case where only one expert is selected from a group of experts---where only one expert $e_{i}$ receives $G_{i}(\mathbf{x}) = 1$ and all the rest zero---makes this partition hard:  each expert specializes on one and only one cell of a partition over $\mathcal{X}$.  We will call this the top-expert MOE.  In that case, $h(\mathbf{x}; n)$ is simply a piecewise function that applies the experts $e_{i}(\mathbf{x})$\footnote{Here we treat the experts as just functions mapping features $\mathbf{x}$ to targets $\hat{y}$ such that $e_{i}(\mathbf{x}) = \hat{y}$.} on $\mathbf{x}$ depending on what member of the partition $Z_{i} \in \{Z_{1}, \dots, Z_{n}\}$ of $\mathcal{X}$ that $\mathbf{x}$ happens to belong in:

\begin{equation}
     h(\mathbf{x}; n) = \begin{cases} 
      e_{1}(\mathbf{x}) & \mathbf{x} \in Z_{1} \\
      e_{2}(\mathbf{x}) & \mathbf{x} \in Z_{2} \\
      \vdots & \vdots \\
      e_{n}(\mathbf{x}) & \mathbf{x} \in Z_{n} 
   \end{cases}
\end{equation}

\noindent Here expert $e_{1}$ only provides advice on cell $Z_{1}$, $e_{2}$ on cell $Z_{2}$, and so on.  The intuition behind MOE is familiar to consumers of medicine; when presented with symptoms, we will often go to the doctor we think is best specialized in diagnosing and treating illnesses that typically present those symptoms.  If we happen to choose only one doctor for clinical advice, we employ the top-expert MOE, and if the doctors are all broadly similar in how they approach problems---say by receiving standard medical training instead of alternative medicine---then they are homogenous.\footnote{Of course, this analogy breaks down when we consider how MOEs are trained:  namely, we don't learn which doctor to use at the same time they receive medical training.}

MOEs and ensembles, while both aggregation techniques, have important differences.  Those differences consist of 1) how they are trained and 2) how they aggregate predictions.

In the training regime, ensembles train their models separately and independently from their aggregation scheme.  For example in bagging \citep{breiman1996bagging}, models are trained separately on random subsets of data drawn from $\mathcal{D}$ and only combined at inference time with their aggregation scheme.  Boosting \citep{schapire1990strength, freund1996experiments} modifies this regime by training models sequentially on the whole data but weighting data samples by the errors of the previous model in the sequence to gradually scale up better predictors.  In contrast, the experts and the gating function are all trained simultaneously in MOE; the gating function learns the relevant soft partition while the experts specialize in their particular region of data.  This is done with either gradient descent or the expectation maximization algorithm.  Unlike ensembles, this means the aggregation scheme is not done separately in inference but is crucial to the training process because of the tight connection between the gating function and the experts.

It is important to emphasize that the aggregation schemes between ensembles and MOEs differ in the relevance of the sample features for aggregating predictions.  In ensembles, those features are completely ignored.  For example, averaging schemes use weights that are explicitly not a function of the sample---though they may be informed by the training data.  MOEs, however, use a gating function that explicitly routes samples to experts based on the sample features.  Most weighting schemes are essentially conditional probabilities $p(e | \mathbf{x})$ that up-weight the relevant experts and down-weight all the others based on the observed sample and what partition its features fall in.  So aggregation proceeds very differently between ensembles and MOEs.

An extremely influential ensemble scheme is BMA, where all possible predictors of a certain class are considered by requiring the weights $w_{i}$ to be probabilities that reflect the posterior distribution of the model parameters given the data observed.  This ensemble method has important theoretical guarantees, and we turn to discussing it now.
\section{Bayesian Optimality}\label{sec:bayes}

BMA receives its name from the application of Bayes rule to statistical and machine learning model prediction.  Recall that Bayes rule states that if $\theta_{1}, \theta_{2}, \dots$ is a partition of model parameters, then the conditional probability of an element of that partition $\theta_{i}$ given some observed data $\mathcal{D}$ is the ratio of the likelihood times the prior probability and the marginal probability of the data:\footnote{When we deal with an uncountably infinite partition of parameters $\theta$, this becomes the integral $p(y | \mathbf{x}, \mathcal{D}) = \int p(y | \mathbf{x}, \theta)p(\theta | \mathcal{D}) d\theta$.}

\begin{equation}
    p(\theta_{i} | \mathcal{D}) = \frac{p(\mathcal{D} | \theta_{i})p(\theta_{i})}{p(\mathcal{D})} = \frac{p(\mathcal{D} | \theta_{i})p(\theta_{i})}{\underset{j=1}{\overset{\infty}{\sum}}p(\mathcal{D} | \theta_{j})p(\theta_{j})}
\end{equation}

\noindent This is called the posterior probability, and it is applied in the \textit{posterior predictive distribution} $p(y | \mathbf{x}, \mathcal{D})$.  The key idea is that predicting some target $y$ from features $\mathbf{x}$ and some previously observed data can be had by applying the law of total probability with respect to the model partition $\theta_{1}, \theta_{2}, \dots$ and weighting the model likelihoods by the posterior of the model given the data:

\begin{equation}
    p(y | \mathbf{x}, \mathcal{D}) = \underset{i=1}{\overset{\infty}{\sum}} p(y | \mathbf{x}, \theta_{i})p(\theta_{i} | \mathcal{D})
\end{equation}

\noindent In essence, this states that if we average over all possible models by how likely those models are correct given the data we have observed, then we can compute a posterior of the target given the observed data.\footnote{Crucially, we arrive at the posterior predictive distribution by way of assuming conditional independence of the sample $\mathbf{x}$ and the data $\mathcal{D}$ given the parameters $\theta$ and the conditional independence of the parameters $\theta$ and the sample $\mathbf{x}$ given the data $\mathcal{D}$.}  This has several desirable properties.

First, it allows us to form predictions with a natural regularizer.  Regularization is a technique common in machine learning to combat overfitting.  Overfitting is where a machine learning algorithm essentially does well on training data while failing to predict hold-out test data; the model ``memorizes'' the training data instead of truly learning the relevant ``inductive patterns in the data''.  An explanation for this behavior comes from the Probably, Approximately Correct (PAC) learning framework \citep{kearns1994introduction}:  models that overfit are too ``complex'' for the data.\footnote{It turns out that ``complexity'' is a red-herring for the guarantees of PAC-learning as \citet{herrmann2020pac} demonstrates.  See \citet{sterkenburg2023statistical} for further discussion and defense of PAC-learning.}  Consequently, model complexity is penalized by an additional term in the loss function during training.  This is called regularization, and the additional term in the loss function that penalizes complexity is called a regularizer.  In the case of BMA, the posterior $p(\theta | \mathcal{D})$ builds in a natural regularizer with the prior $p(\theta)$.  So BMA naturally provides better generalization and leads to ``simpler'' models over other aggregation techniques.

Second, BMA produces models that have good decision-theoretic properties.  We consider our loss function as capturing the relative dis-utility of making decisions, and a machine learning model as providing a decision rule (here we treat predictions as a type of action).  Then if we suppose there is a ``true'' parameter $\theta$ for generating the data $\mathcal{D}$, then we say the frequentist risk of a model is the likelihood of the data given the true parameter weighted expected value of that model's loss.  For example, when predicting the outcome of a sequence of coin tosses, the frequentist risk would be how well our model does on some loss for those tosses weighted by how likely each toss is given the bias of the coin.  So supposing us to be realists about chances, the frequentist risk captures how well a machine learning model makes decisions relative to the true chances.  This notion of risk has a tight connection to BMA.  A BMA model is a \textit{Bayesian estimator} in the sense that it minimizes the Bayes risk with respect to the posterior $p(\theta | \mathcal{D})$:  the Bayes risk of a model with prior $\rho (\theta)$ is simply the $\rho (\theta)$ weighted expected value of the frequentist risk.  Instead of assuming a ``true'' parameter, we propose a prior probability distribution for the parameter and average out that prior across the frequentist risk.  Suppose we are unsure what the true bias of a coin is when observing a sequence of coin tosses.  Then we can consider the Bayes risk as putting a prior distribution over the bias and weighting the frequentist risk by that prior.  If our model of the coin toss minimizes that risk, then we say it is a Bayesian estimator for the aforementioned prior.  Consequently, it should be clear that a BMA model is a Bayes estimator concerning the posterior $p(\theta | \mathcal{D})$ because of how it is defined.  

The so-called complete class theorems are an important property connecting the frequentist risk and the Bayes risk.  These theorems state that Bayesian estimators form a complete class in the sense that any machine learning model that does better on the frequentist risk than any other model is a Bayesian estimator \citep{wald1947essentially, robert2007bayesian, murphy2022probabilistic}.\footnote{More precisely, one model is said to dominate another model if the frequentist risk for the former is less than or equal to the latter across all possible true parameters $\theta$; we say a model is admissible if no other model is better than it, i.e. no other model strictly dominates it.  Then the complete class theorem states that any admissible model is a Bayesian estimator with respect to some, possibly improper, prior \citep{wald1947essentially}.}  This means that there will be some Bayesian prior we can average over to equal the ``correct'' model of the data that does best on the loss; the hope is that we can then find that prior through our posterior by conditioning on the data.  So, ideally, a BMA with enough data and non-perverse prior will be the Bayesian estimator that does as well as one can do on the frequentist risk.  This should show up in terms of minimizing the loss on the hold-out test set.

With both of these desirable properties, it is then surprising that BMA is often outperformed by a MOE.

\begin{figure}
    \centering
    \begin{subfigure}{0.45\textwidth}
        \includegraphics[width=\textwidth]{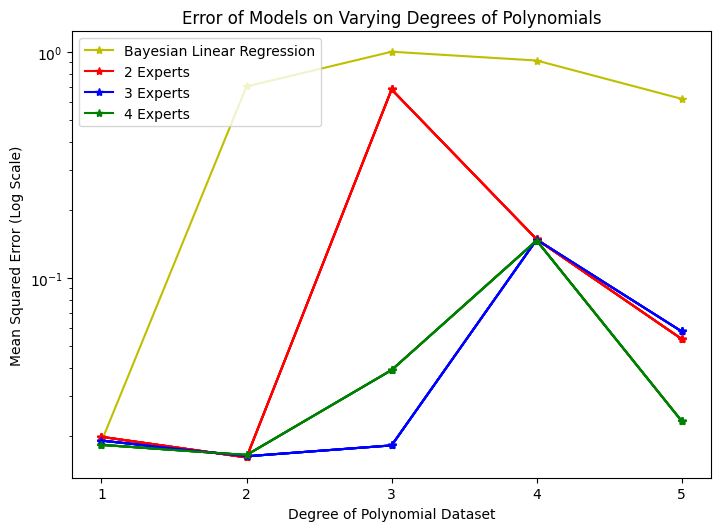}
        \caption{Mean Squared Error}
        \label{fig:bayes-MOE-mse}
    \end{subfigure}
    \begin{subfigure}{0.45\textwidth}
        \includegraphics[width=\textwidth]{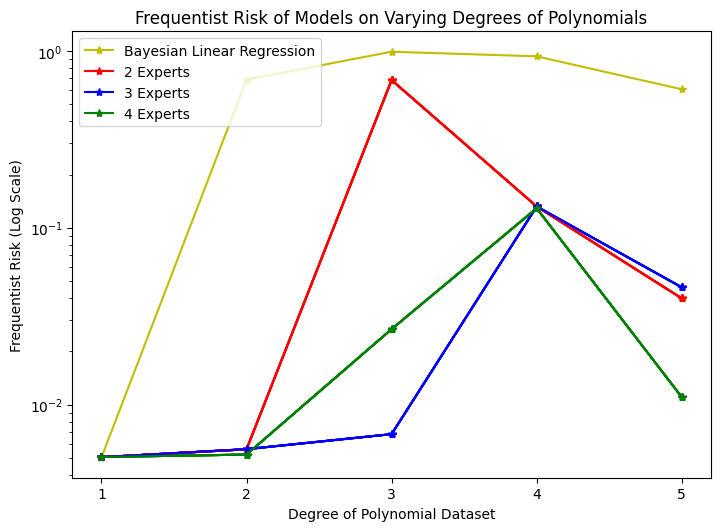}
        \caption{Frequentist Risk}
        \label{fig:bayes-MOE-risk}
    \end{subfigure}
    \caption{Plots of the mean squared error and frequentist risk for Bayesian Linear Regressions and MOE with $[2,3,4]$ expert linear regressions on polynomial datasets of degrees $[1,2,3,4,5]$.  Lower mean squared error and frequentist risk is better.}
    \label{fig:bayes-MOE-mse-risk}
\end{figure}

In figure \ref{fig:bayes-MOE-mse-risk}, we show experiments involving a type of BMA and similar MOEs.  The former is a Bayesian linear regression, which averages across a posterior $\mathcal{N}(\theta | \mu, \Sigma)$ all linear regressions given in equation \ref{eq:linear-regression}, and the latter is a MOE involving two or more linear regressions as found in equation \ref{eq:linear-regression} (details can be found in appendix two).  Both models were trained on identical training data and evaluated on the same hold-out test data.  The data consists of polynomials with some added normally distributed noise.  Figure \ref{fig:bayes-MOE-mse} shows the performance of these models on the mean squared error on hold-out test data, where lower is better; all mixture models match or exceed the performance of the Bayesian linear regression---with the greatest gap between the larger number of experts on higher degree polynomials.  Furthermore, since we control the data generation, we can directly calculate the frequentist risk.  Again, the MOEs perform best here as seen in figure \ref{fig:bayes-MOE-risk}, with the results closely tracking the mean squared error where lower is better.  This collides with the expected hope that we should find the BMA model to do better here.

One might think that these results are due to the MOEs employing more sophisticated experts, such as polynomial regressions, or a non-probabilistic aggregation scheme but that is not the case.  Both the Bayesian linear regression and the MOEs average over linear regressions, i.e. the MOEs are homogenous:  each expert is accounted for in the Bayesian linear regression, and those experts receive some weighted probability according to the posterior.  If anything, the MOEs employ \textit{fewer} experts since a BMA averages across all possible expert configurations while the most we consider in our experiments are four experts.  Furthermore, the averaging scheme employed here is not all that different; both the Bayesian linear regression and the MOEs weigh their subcomponent models by probabilities.  In fact, the Bayesian linear regression does things the right way by weighting the models by a data-driven posterior.  So the difference between the two cannot be due to better experts or a non-probabilistic approach to aggregation.

The ability of MOE to outperform ensembles like BMA has led them to be an important component of contemporary machine learning techniques.  For example, \citet{yuksel2012twenty} cite a 2008 survey of then-important machine learning methods and argue that MOE outperforms all of them \citep[1178]{yuksel2012twenty}.  They in particular cite the advantages a MOE model has over the popular ensemble method of boosting \citep[1187--88]{yuksel2012twenty}.  Furthermore, \citet{masoudnia2014mixture} argue that MOE consistently outperforms popular ensemble techniques like bagging and boosting \citep[286--287]{masoudnia2014mixture}.  This is despite the challenges MOE faces in learning the right partition and gating function (see section 3.3 of \citet{masoudnia2014mixture} for a discussion circa 2014).  Those challenges have increasingly been overcome through sparsity, i.e. only selecting the top-$k$ experts when making predictions, and noisy gating, i.e. injecting noise during the training process to force the gating function to use all experts \citep{shazeer2017outrageously}.  This has led to MOEs being an increasingly important component of large language models, like OpenAI's GPT-4 \citep{betts2023gpt4} and Mistral's Mixtral \citep{jiang2024mixtral}.

Two explanations have been given for the superiority of MOE over ensemble methods like BMA.  First, MOEs work well on data generated by oxymoronic one-to-many functions, i.e. ``functions'' that can map a single input to multiple outputs, whereas traditional models and ensembles cannot learn this data \citep[454]{murphy2022probabilistic}.  However, this explanation fails in the experiments described above:  all of the datasets were generated by proper functions that pass the vertical line test.  Second, MOEs supposedly work well due to experts learning data that is negatively correlated \citep[287]{masoudnia2014mixture}.  While this is true for some data, it is not always the case, as can be seen in the aforementioned experiments on degree three polynomials where different elements of the data are either positively correlated or not correlated at all.  So both of these explanations seem unnecessary for explaining why MOE does well.

Recapping, an important class of ensemble models are Bayesian ones.  These models have attractive theoretical features such as regularization and accuracy considerations.  But these models are often outperformed by MOEs as demonstrated by some experiments.  This has led to a general adoption of MOEs in machine learning for hard problems over ensemble schemes, such as in large language models.  Two reasons for this superiority are that MOEs can fit data generated by one-to-many ``functions'' and that MOEs train their experts on negatively correlated data.  But better performance can be found on proper functions and the partition learned need not involve negative correlations.  So another explanation should be given, which we turn to now.
\section{The Functional Capacity of Models}\label{sec:capacity}

A plausible, intuitive hypothesis for why MOEs outperform ensemble techniques like BMA is that MOEs have greater functional capacity than ensembles.  That is, MOEs characterize hypothesis classes that have a richer set of labeling schemes than ensembles like BMA, and so the improved performance relative to provably ``better'' methods like BMA is due to BMA being restricted to a more impoverished model class.  This shows up in the experiments in section \ref{sec:bayes} where we see that the various MOEs continue to do well on mean squared error loss as the degree of the polynomial data scales.  The Bayesian is in some sense limited by only considering linear regressions, while the MOEs seem able to stitch together multiple functions---despite the MOEs, like the BMA, only using linear regressions.

To make this hypothesis exact, we need to do two things:  first, we need to specify what is a good explication of functional capacity, and second, we need to show how, in principle, at least some MOEs have a greater functional capacity than a BMA.

The intuitive idea of the functional capacity of a model is supposed to be something like the set of functions the model can learn.  What counts as a relevant function depends on the problem the machine learning model solves.  In supervised learning, those problems are sorted into the buckets of regression and classification, which correspond to predicting continuous targets or discrete targets respectively.  With classification, the prediction problem is one of separating data into sets identified with the target, and the complexity of the problem is in some sense dependent on the data being used---predicting the label of a picture from raw pixels is harder than predicting from a higher-level feature like the presence or absence of an animal.  Measuring how well we can separate data is the core intuition behind the Vapnik-Chervonenkis dimension (VC dimension).  A more sophisticated model with higher functional capacity will be one that can better sort more complex data into classes; VC dimension gives us an exact measure of that capacity in terms of the cardinality of the data.  This makes it a good explication of functional capacity because it specifies the type of functions a model can fit by one important sense of the complexity of the data in the classification problem at hand.

To define VC dimension, we need to define what it means to separate data, i.e. to \textit{shatter} a dataset.  Our context is binary classification, where machine learning models are treated as functions that map between features $\mathcal{X}$ and the binary labels $\{0,1\}$.  For example, we might be interested in classifying whether a picture is of a cat or not.  We then group our binary classifiers into sets $\mathcal{H}$ and specify that $\mathcal{H}$ shatters some dataset $X$ if for any arbitrary labeling of $X$ there is a classifier in $\mathcal{H}$ that correctly classifies that labeling:

\begin{definition}
    Let $\mathcal{H}$ be a set of binary classifiers $h: \mathcal{X} \to \{0,1\}$.  Given a set of points $X = \{x^{(1)}, \dots, x^{(m)}\}$ where $x^{(i)} \in \mathcal{X}$, we say that $\mathcal{H}$ shatters $X$ if for any labels $Y = \{y^{(1)}, \dots, y^{(m)}\}$ where $y^{(i)} \in \{1, 0\}$, there exists $h \in \mathcal{H}$ such that $h(x^{(i)}) = y^{(i)}$ for $i = 1, \dots, m$.
\end{definition}

\begin{figure}[t]
    \centering
    \begin{subfigure}{0.45\textwidth}
        \centering
        \begin{tikzpicture}
            \node[label=below:{\scriptsize (0,0)}, circle, fill=blue, inner sep=2pt] (a) at (0,0) {};
            \node[label=above:{\scriptsize (1,1)}, circle, fill=blue, inner sep=2pt] (b) at (1,1) {};
            \node[label=below:{\scriptsize (2,0)}, circle, fill=red, inner sep=2pt] (c) at (2,0) {};
    
            \draw[thick, dashed] (0.5,-0.5) -- (2.25,1.25);
        \end{tikzpicture}
        \caption{Shattering 3 points}
        \label{fig:shatter-3}
    \end{subfigure}
    \begin{subfigure}{0.45\textwidth}
        \centering
        \begin{tikzpicture}
            \node[label=below:{\scriptsize (0,0)}, circle, fill=red, inner sep=2pt] (d) at (0,0) {};
            \node[label=above:{\scriptsize (1,1)}, circle, fill=red, inner sep=2pt] (e) at (1,1) {};
            \node[label=above:{\scriptsize (0,1)}, circle, fill=blue, inner sep=2pt] (f) at (0,1) {};
            \node[label=below:{\scriptsize (1,0)}, circle, fill=blue, inner sep=2pt] (g) at (1,0) {};
    
            \draw[thick, dashed] (-0.5,0.5) -- (1.5,0.5);
        \end{tikzpicture}
        \caption{Not shattering 4 points}
        \label{fig:shatter-4}
    \end{subfigure}
    \caption{An example of shattering and failing to shatter.  Figure \ref{fig:shatter-3} shows a set of points that can be shattered by the set of linear classifiers because any arbitrary labeling, i.e. which points are assigned blue or red, can be correctly classified by at least some linear classifier, i.e. we can draw a line separating the two labels without any mistakes.  Figure \ref{fig:shatter-4} shows a set of points that cannot be shattered by that set since no line can separate this particular coloring without any errors.}
    \label{fig:shattering}
\end{figure}
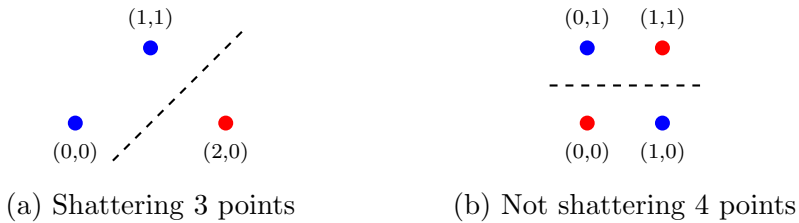

\noindent Suppose our set of models are all linear classifiers, i.e. they classify by drawing lines to separate data.  Then as can be seen in figure \ref{fig:shattering}, this class can shatter a given set of three points in $\mathbb{R}^{2}$ seen in figure \ref{fig:shatter-3}, but there is no set of four points that it can shatter as seen in figure \ref{fig:shatter-4}.  It shatters the former because regardless of how we label those three points, some line can separate them, which implies there is some linear classifier that can make a completely correct and error-free identification of the targets.  It fails to shatter the latter because those four points have a labeling that no line can cleanly separate error-free.

The VC dimension of a class $\mathcal{H}$ is then the cardinal size of the largest set of data, with respect to some space of possible data like $\mathbb{R}$, that that class can shatter:

\begin{definition}
    The Vapnik-Chervonenkis dimension of a set of binary classifiers $VCD(\mathcal{H})$ defined over instance space $\mathcal{X}$ is the cardinality of the largest finite subset of $\mathcal{X}$ that $\mathcal{H}$ can shatter.  If arbitrarily large subsets can be shattered, then $VCD(\mathcal{H}) := \infty$.
\end{definition}

\noindent One can understand VC dimension of size $m$ with a two-player game.  Player one starts with a set of binary classifiers.  A round begins by her picking a set of data points of size $m$ from the instance space (they can be any points she chooses); player two then assigns any labeling he wants to those points.  If player one has a binary classifier that correctly labels all those points, she wins the round, and player two picks another set of labels; if she doesn't, then player two wins the round, and player one then must choose another set of points of size $m$.  Should player two exhaust all of the labels for player one's chosen set, then player one has successfully shattered the set and wins the game, and the VC dimension is at least size $m$.  But if player one cannot find any dataset after complete play, player two wins the game, and the VC dimension less than size $m$.  The exact VC dimension will then be the largest set that player one can win.  For example, playing this game with linear classifiers in the Cartesian plane will result in player one winning with a set of three points and player two winning when the number of points grows to four.  So the VC dimension of the set of linear classifiers concerning $\mathbb{R}^{2}$ will be exactly three.

It should be noted that VC dimension is not correlated with parameter count:  one can have a high VC dimension with a low parameter count.  For example, the set of models given by $I(\sin \alpha \cdot \mathbf{x})$, where $I(x)$ is the indicator function that assumes $1$ if $x > 0$ and $0$ otherwise, has only one parameter $\alpha \in \mathbb{R}$ but an infinite VC dimension \citep[237]{hastie2009elements}.  This makes VC dimension a better characterization of model functional capacity than raw parameter count since the above model intuitively is more complex than a linear classifier---even though the linear classifier has more parameters.

Importantly, ensembles have a maximum VC dimension given by the number of ensemble members and the base VC dimension of the class of members used in the set of ensemble hypotheses.  Let $L(\mathcal{H}, n)$ be the set of binary classifiers that are linear combinations of hypotheses drawn from binary classifiers $\mathcal{H}$, i.e. $L(\mathcal{H}, n) = \{x \mapsto \sgn(\sum^{n}_{i=1}w_{i}f_{i}(x)): \mathbf{w} \in \mathbb{R}^{n}, \forall i, f_{i} \in \mathcal{H}\}$.  By definition, every ensemble will be in this set.  Then it can be shown that if the $VCD(\mathcal{H})=m$ is finite for some finite $m > 3$, then $VCD(L(\mathcal{H}, n)) \leq \overset{\sim}{O}(mn)$, where $\overset{\sim}{O}$ means we ignore any constants or logarithmic terms \citep[109]{shalev2014understanding}.\footnote{More precisely, the result is provided in \citet[109]{shalev2014understanding} with Lemma 10.3 as:

\begin{equation*}
    VCD(L(\mathcal{H}, n)) \leq n(VCD(\mathcal{H}) + 1)(3 \log n(VCD(\mathcal{H} + 1)) + 2)
\end{equation*}}  This is effectively to say that the VC dimension of ensembles with $n$ members and whose members have a VC dimension $m$ is no bigger than $nm$.  While these bounds do not technically apply to BMAs, since BMAs consider infinitely many hypotheses, in practice, they do because almost all BMAs are approximated through finite sampling schemes, i.e. Monte Carlo methods.  Crucially, this is an upper bound:  it can be shown in many cases that the VC dimension is significantly less---sometimes it is exactly the same VC dimension as the base class $\mathcal{H}$ (see \citet[113]{shalev2014understanding} exercise 10.4 for examples).  The upshot is that ensembles are bounded above by the number of members and the capacity of those members to separate data.

MOE, however, can be shown to have a higher VC dimension than the experts it employs.  \citet{jiang2000vc} showed that for mixtures of $n$ Bernoulli binary classifiers on $\mathbb{R}$, the VC dimension is exactly $n$, and for logistic regressions or Bernoulli binary classifiers on the instance space of $\mathbb{R}^{d}$, the VC dimension can be bounded below by the number of experts $n$ and from above to the worst case $O(n^{4}d^{2})$ \citep[7]{jiang2000vc}.  These bounds can be tightened on $\mathbb{R}$.  First, we say a set of functions $\mathcal{F}$ whose domain is the reals is closed under translation just in case for any function $f$ in that set and any $a \in \mathbb{R}$, the function $f_{a}(x + a)$ is also in $\mathcal{F}$.  Examples of these sets will be the set of all linear binary classifiers, the set of all polynomial binary classifiers of some positive degree $d$, and the set of all feedforward neural networks whose domains are the reals.\footnote{The set of all linear classifiers and the set of all polynomial classifiers of some positive degree $d$ will be closed under translation simply because the translation will itself just be another linear binary classifier or polynomial binary classifier (we add it to the intercept in the former and to the zeroth term in the polynomial).  For feedforward neural networks, we have to note that a translation on input corresponds to changing the bias of the initial computation layer.  For example, if we have some activation function $f$ and a network with hidden layers where each layer $i$ is given by $h_{i}(x) = f(W_{i}x + \mathbf{b}_{i})$ with $W_{i} \in \mathbb{R}^{d \times 1}$ and $\mathbf{b}_{i} \in \mathbb{R}^{d \times 1}$, then the first computation done in the network becomes:

\begin{align*}
    h_{1}(x + a) & = f(W_{1}(x+a) + \mathbf{b}_{1}) \\
    & = f(W_{1}x + W_{1}a + \mathbf{b}_{1}) \\
    & = f(W_{1}x + \mathbf{b}_{1}^{\prime})
\end{align*}

\noindent where $b_{1}^{\prime}$ is just the new offsetting bias $W_{1}a + \mathbf{b}_{1}$.}  Second, suppose we restrict ourselves to a limiting case involving top-expert MOEs with homogenous experts drawn from a class closed under translation.  In that case, we can show that the VC dimension will have a lower bound that is a product of the number of experts and those experts' VC dimension:

\begin{proposition}\label{prop:moe}
    Let $\mathcal{H}$ be a set of top-expert mixture of experts binary classifiers of $n$ homogenous binary classifier experts drawn from the set $\mathcal{E}$ closed under translation with $VCD(\mathcal{E}) = m$ defined over instance space $\mathbb{R}$.  Then $VCD(\mathcal{H}) \geq nm$ for $\mathbb{R}$.
\end{proposition}

\noindent What this means is that at least some MOEs have a higher VC dimension than BMAs and so greater functional capacity.  They can fit more complex data and so involve fundamentally more complex hypotheses.  The intuition behind proposition \ref{prop:moe} can be seen by observing that a top-expert MOE with homogenous experts is a piecewise function that deploys each expert on exactly one element of a partition of the feature space; such a piecewise function will be able to separate more complex data than any of its components individually.  Likewise, non-top-expert MOEs will form similar piecewise functions, though of possibly less complexity depending on the soft partition learned by the gating function.  The upshot is that with functional capacity understood as VC dimension, we can vouchsafe the earlier intuition that MOEs have greater ``capacity'' and can learn more complex functions than ensembles like BMA.

This means that comparing MOEs to BMAs is an apples-to-oranges comparison.  If MOE and Bayesian models are restricted to sub-models with the same complexity, then MOE will have an inherent advantage on more complex datasets like those observed in section \ref{sec:bayes}.  Provide the Bayesian the right type of models and that advantage will likely disappear.

In summary, we have argued that an intuitive explanation for the superiority of MOE over ensembles like BMA---despite the theoretical reasons for the Bayesian to be advantaged ---is the greater functional capacity of MOEs.  We argued that VC dimension is a good explication of that capacity, and we proved that the VC dimension of some MOEs will be greater than a BMA over models with the same capacity as the MOE.  However, theoretical considerations can only cut so much ice.  The question is whether MOEs, such as those that are not top-experts, can do well on datasets that exactly track greater VC dimensions.  We turn to that now.
\section{Experiments}\label{sec:experiments}

The discussion and results in section \ref{sec:capacity} provide a theoretical reason why MOEs may have greater functional capacity than ensembles like BMAs, but does this show up empirically?  Two problems prevent an immediate acceptance of this fact:  first, proposition \ref{prop:moe} only applies to top-expert MOEs but most MOEs in practice have multiple experts per partition, and second, the proposition applies only if the top-expert MOE is learned when this often may prove difficult or impossible in practice.\footnote{Compare the result with the universal approximation theorems for neural networks \citep{hornik1989multilayer}:  those theorems give something like a bound on the expressivity of neural networks as function approximators but those theorems are no insurance that a given network will in fact approximate the desired function.}  Consequently, we should empirically validate our theoretical results.  We address this worry here by detailing experiments involving datasets that stand as a proxy for the VC dimension of binary classifiers.

\begin{figure}[t]
    \centering
    \includegraphics[scale=0.4]{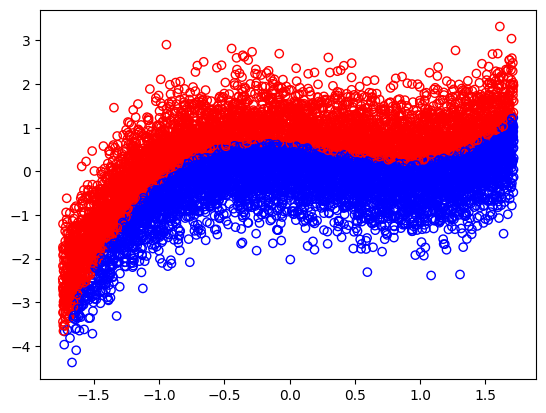}
    \caption{Example of a VC dimension Polynomial Dataset, degree 3.  Red indicates the polynomial binary classifier assigns label $0$ and blue indicates the classifier assigns label $1$.  The resulting correlated VC dimension is $\binom{5}{2} = 10$.}
    \label{fig:ex-vc-dim}
\end{figure}

Our method for building datasets with a correlated VC dimension takes advantage of the fact that polynomial binary classifiers of degree $m$ over instance space $\mathbb{R}^{d}$ have a VC dimension of $\binom{d+m}{d}$ \citep[57]{shalev2014understanding}.  We can then use those binary classifiers to generate datasets that in fact will be classifiable by machine learning models with the appropriate VC dimension.  An example dataset can be seen in figure \ref{fig:ex-vc-dim}.  We build a polynomial binary classifier of degree $m$ with coefficients $[\theta_{m}, \dots, \theta_{0}]$, and then we generate points in $\mathbb{R}^{2}$ around that the polynomial with a normal distribution.  Points that lie above the polynomial are assigned one label and those below another label.  We say this dataset has a correlated VC dimension of $\binom{2+m}{2}$ since we can achieve near-perfect accuracy by fitting a logistic regression with a design matrix that is a degree $m$ polynomial.

Our chosen binary classifiers for the experiments are logistic regressions.  A logistic regression is a binary linear classifier with likelihood $f_{\sigma}(y | \mathbf{x}, \theta)$ where $\theta = [\theta_{0}, \dots, \theta_{d}]^{\intercal}$ and $\mathbf{x} = [1, x_{1}, \dots, x_{d}]^{\intercal}$ such that:

\begin{equation}
    f_{\sigma}(y | \mathbf{x}, \theta) = \text{Ber}(y | \sigma(\mathbf{x}^{\intercal}\theta))
\end{equation}

\noindent Here Ber is the Bernoulli distribution and $\sigma$ is the logistic sigmoid function $\sigma(x) = \frac{1}{1 + \exp{(-x)}}$.  Our MOEs consist of two to four experts that are logistic regressions that employ a top-2 expert configuration, and we choose a BMA that averages over logistic regressions.  Unfortunately, there are no analytically computable posteriors $p(\theta | \mathcal{D})$ for logistic regressions, so we have to approximate it for the BMA (see appendix two for details).

\begin{figure}[t]
    \centering
    \begin{subfigure}{0.45\textwidth}
        \includegraphics[width=\textwidth]{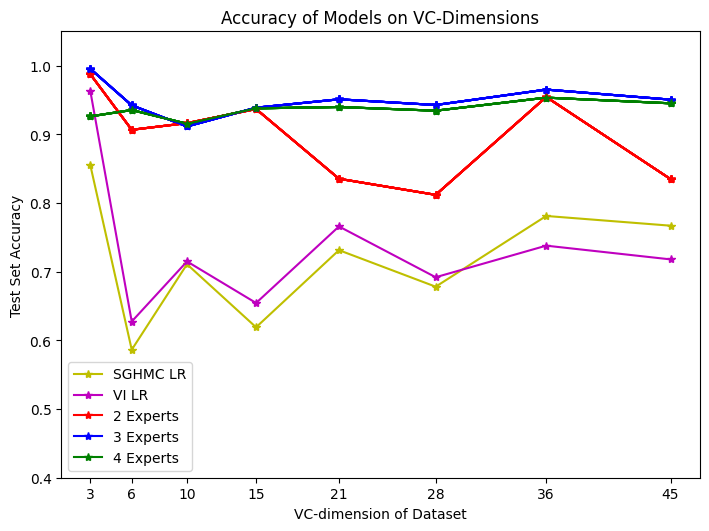}
        \caption{Accuracy}
        \label{fig:vc-dim-acc}
    \end{subfigure}
    \begin{subfigure}{0.45\textwidth}
        \includegraphics[width=\textwidth]{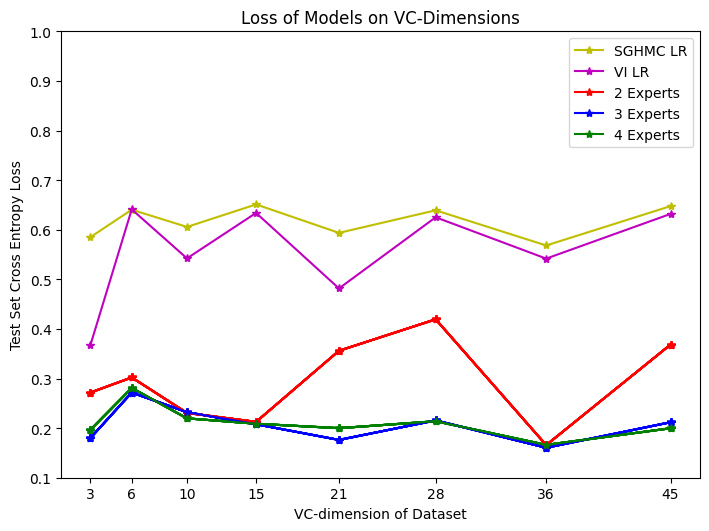}
        \caption{Loss}
        \label{fig:vc-dim-loss}
    \end{subfigure}
    \caption{The hold-out test set accuracy and loss of SGHMC and VI logistic regression (LR) BMAs and logistic regression MOEs.  For accuracy in figure \ref{fig:vc-dim-acc}, higher is better; for loss in figure \ref{fig:vc-dim-loss}, lower is better.  We see that as the VC dimension increases, the accuracy and loss of the BMAs falls off, while the accuracy of the MOEs says relatively constant, with some degradation in the two-expert model.}
    \label{fig:vc-dim-models}
\end{figure}

All models were trained to minimize the cross-entropy loss function using stochastic gradient descent on a training set, and we then evaluated them on a hold-out test set.  The results can be seen in figure \ref{fig:vc-dim-models}.

The experiments show that at a correlated VC dimension of $3$, the Bayesian models perform roughly on par with the MOEs, but then there is considerable separation as the VC dimension of the dataset climbs.  Figure \ref{fig:vc-dim-acc} shows that BMAs consistently underperform on test-set accuracy, only doing modestly better than chance; however, the MOEs perform well as the VC dimension continues to climb, though the MOE with two experts shows degraded accuracy relative to the other models.  In figure \ref{fig:vc-dim-loss}, we see the same trends on test-set accuracy reflected in the cross-entropy loss:  the Bayesian models approach the loss on hold-out test data as the MOEs, but then underperform as the VC dimension climbs.  Again, the two-expert MOE model shows degradation more so than the higher expert counts.

These results validate empirically the theoretical results from section \ref{sec:capacity}.  Note that the bound proved in proposition \ref{prop:moe} is a lower bound for top-experts, so the VC dimension of the MOEs may be higher.  The upshot is that we have both theoretical and empirical reasons to suggest that our explanation for the superiority MOEs over ensembles like BMA is correct:  MOEs can simply learn more expressive hypotheses than averaging schemes like BMAs.
\section{Mixtures of Experts as Abductive Modeling}\label{sec:abduction}

The argument from the previous sections has been that MOEs outperform averaging schemes like the Bayesian ones because they avail themselves of a richer hypothesis space.  What is very interesting about this is that they find themselves in that bigger hypothesis space by way of simpler hypotheses; the experts, as noted, have a lower functional capacity than the MOE as a whole.  Through the gating function, the feature space is soft-partitioned into cells that expert models can then specialize on.  Importantly, this allows a virtuous learning process that enables experts to master relatively narrow domains while ensuring that collectively the experts build a more sophisticated, broader picture of the problem at hand.  We argue in this section that this process is a form of \textit{abduction} in the Peircian sense of hypothesis construction.  This explains the fundamental difference between MOE and BMA:  the former combines abduction and induction while the latter only uses induction.

A big assumption of the Bayesian approach to inductive reasoning is that the hypothesis space must be given.  We have an algebra (or sigma-algebra) that captures in some sense all of the questions that we want to have answered in our inquiry and evidence we could bear on those questions.  Then we proceed in our inquiry by considering some partition that is a subset of that algebra that characterizes the relevant scientific hypotheses and applying Bayes rule on new evidence as it comes in to update ourselves on our credences in the members of that partition.  The assumption about the hypothesis space shows up twice here:  first, in the granted algebra that characterizes inquiry, and second, in the selected hypothesis partition used when applying Bayes rule.  We see both assumptions play a crucial role in BMA.  The initial algebra is the possible model parameters and the range of values the feature and target random variables may assume, and the hypothesis partition is the range of possible model parameters assumed.  With this fixed, no further hypothesis or evidence may be admitted into the averaging scheme.  So BMA perfectly embodies the Bayesian approach---an approach that is fundamentally an inductive one.

By an inductive approach, we mean an ampliative inference from evidence to hypotheses.  For the Bayesian, this can be thought of as a type of diachronic coherence between our current partial beliefs and our future partial beliefs; we change our beliefs in such a way as to avoid sure losses when gambling on those beliefs \citep[132--133]{zabell2005symmetry}.  In that sense, BMA is induction \textit{par excellence}---we aim to make predictions of new data based on what we have observed from old data in a probabilistic manner that guarantees we do as best we can by a loss function that captures the modeler's subjective utility function on the prediction problem.  The main assumption we make is that some class of models best parameterizes the predictive problem, and so we learn about those models based on the data we have observed in a diachronically coherent manner.  Then when we make our predictions, we factor what we have learned about those models into our predictions.

While MOE certainly does induction too, there is an additional element not present in the Bayesian approach.  MOE partitions the input feature space and then through the gating function routes predictive samples to experts based on where they fall in that partition.  This is termed a divide-and-conquer strategy or conditional computation since the MOE learns how best to divide a problem into sub-problems that can more easily be addressed by expert models.  In contrast, BMAs learn no such partition nor do they break down the problem into more easily addressable sub-problems.  This divide-and-conquer approach is chiefly responsible for MOE's expressive advantage over ensemble methods.

We argue that the MOE approach to the problem of prediction in machine learning is exactly an approach germane to abductive inference or hypothesis construction. 

What we mean here by abductive inference or abduction is not inference to the best explanation defended in \cite{douven2022art}.  Instead, we mean an older idea defined by C. S. Peirce about guessing or conjecturing hypotheses:

\begin{quote}
    Accepting the conclusion that an explanation is needed when facts contrary to what we should expect emerge, it follows that the explanation be such a proposition as would lead to the prediction of the observed facts, either as necessary consequences or at least as very probable under the circumstances.  A hypothesis, then, has to be adopted, which is likely in itself, and renders the facts likely.  This step of adopting a hypothesis as being suggested by the facts, is what I call \textit{abduction} \citep[94--95]{peirce1901logic}.
\end{quote}

\noindent The idea is not to find what hypothesis best explains the facts but to find a hypothesis period to be considered for testing.  Peirce thinks that this is perhaps the most important element of scientific inquiry, and he discusses several principles to be applied when conjecturing hypotheses.  Among those to consider is the economies of research---the resources spent finding hypotheses and testing them---which are important because we should expect our candidate hypotheses to break down and our time and resources in inquiry limited.  He proposes several qualities to guide hypothesis selection \citep[109]{peirce1901logic}.  One in the current context stands out.  He calls this ``Caution'' and what he means is adopting a strategy of hypothesis formation that breaks down the problem space into regions that allow for groups of hypotheses to be quickly tested and discarded by their sub-components to converge at the best candidate in the most expeditious manner.  To illustrate this idea, he gives as an example the two-player game of guessing twenty questions, where one player thinks of some object and the other player has twenty yes-or-no questions to divine the object the first player has in mind.  Peirce argues that a good player chooses questions that sequentially equally divide the possible object space.  He analogizes cautionary hypothesis construction to such a procedure where ``The secret of the business lies in the caution which breaks a hypothesis up into its smallest logical components, and only risks one of them at a time'' \citep[109]{peirce1901logic}.  While there is some ambiguity about what Peirce means here, one plausible interpretation is that hypothesis construction should proceed with an eye toward the logical parts of the hypothesis and how they can be structured in a way to facilitate a divide-and-conquer search when testing the hypothesis.  We select a hypothesis based on how fruitfully its components divide up the testing space.  That selection strategy functions as an integral component in abductive reasoning.

The same divide-and-conquer strategy goes on in MOE at the time of inference where a complex sample is routed to a region where a smaller, simpler component can be used to make a prediction.  A MOE model as a hypothesis breaks down the prediction problem into sub-problems with the learned partition, and then it deploys the relevant experts to make its predictions.  Similarly, it also applies this divide-and-conquer strategy in reverse when constructing the MOE model, since the structure of the model architecture and training environment ensures that a MOE learns a good way to divide up the input space to make it tractable for the simpler, expert predictors.  In both cases, a new hypothesis is constructed with an eye toward its simpler logical components and the input space---much akin to the strategy Peirce gave for picking cautionary, economical hypotheses.

How new hypotheses are constructed can be easily observed by the limiting case of top-expert MOEs.  There the new hypothesis given by the MOE model is just a piecewise function of the expert models based on the partition the gating function learns.  This is a more complex hypothesis, as we argued in section \ref{sec:capacity}, since it can fit more complex data than the individual components.  And crucially, it is not simply a weighted combination of the expert models due to the gating function:  models are only applied based on the input and where that input falls in the partition learned by the gating function.  A piecewise function is typically not just a replication or Boolean of its parts.  So at least in the case of the top-expert MOE, something truly new is produced.

And similar to Peirce's guessing game, we can iterate the partitioning strategy to sequentially divide-and-conquer the prediction problem---leading to even more expressive hypotheses.  Hierarchical Mixtures of Experts \citep{jordan1994hierarchical} does exactly this by applying gating functions sequentially to form a tree structure where the experts act as leaves; each gate partitions the cell selected by the previous gate eventually terminating in an application of the relevant experts.  This process is akin to asking questions that rule out more and more alternatives until the best hypothesis is lighted upon.

One thing that should be noted is that the abductive process used in MOEs has an element of induction.  Namely, the gating function and experts are learned simultaneously by how well the MOE as a whole predicts the training data.  This suggests that the process of hypothesis formation is intimately tied to having a positive feedback loop with how well the hypothesis can be used experimentally; forming hypotheses requires putting hypotheses to the test and using the evidence from those tests to guide in further hypothesis selection.  This is a point that Peirce repeatedly hits upon, and we see that this virtuous process plays out in the case of MOEs.

In short, MOEs embody a form of hypothesis construction.  That hypothesis construction can be seen in the limiting case where we have a top-expert MOE with homogenous experts; we have a new hypothesis constructed as a piecewise function of simpler hypotheses that has provably more functional capacity than those original hypotheses.  Importantly, the construction of new hypotheses proceeds in combination with inductive reasoning where the gating function and experts are learned at the same time by how well they predict the training data.  This enables beneficial learning due to how hypothesis construction can be guided by inductive feedback on the training data---leading to better hypotheses through the pressure of gradient descent or another learning algorithm.
\section{Conclusion}

In this paper, we considered the question of why MOE methods perform better than ensemble methods such as BMA---even though BMA has certain optimality guarantees.  We argued that in a limiting case, MOE has a greater functional capacity, in the precise sense of VC dimension, than ensembles like BMAs.  We then demonstrated with a series of experiments that this limiting case seems to track the more usual cases machine learning researchers encounter.  The philosophical upshot is that MOEs seem to employ a type of abductive reasoning in the Peircian sense of hypothesis construction; they use a divide-and-conquer strategy to compose new, more complex hypotheses out of simpler hypotheses in an economical manner.  This explains the discrepancy between Bayesian methods and MOEs because the former are fundamentally just inductive methods while the latter can in some sense be said to construct the hypothesis spaces that the former rely upon.

We conclude by discussing two open questions:  do these results imply that MOEs are just better than BMAs and can the greater functional capacity of MOEs pose inductive problems?

Returning to the question about the superiority of MOE over BMA, one might conclude from this discussion that MOE is the superior algorithm in machine learning over BMA.  We think this is premature.  We have argued that the superiority of MOE over BMA is due to the former applying a form of abductive reasoning in combination with inductive reasoning while BMA is inductive only.  However, the inductive methods in BMA are at least as good as the inductive methods used in MOE---if not better; this means that MOE could likely be improved by incorporating BMA into the training process through which gates and experts are learned.  Instead of using a single or a finite subset of experts who specialize in soft partitions of the feature space, there could be a benefit of employing BMAs here.  Furthermore, a BMA gating function might also improve the learning process.  The upshot is that the abductive process captured in MOEs could be further improved by incorporating a superior inductive process in MOE model formation.

Lastly, we have argued that the superior performance of MOE over BMAs is fundamentally due to BMAs \textit{underfitting} the data.  The greater VC dimension of MOEs allows them to separate more complex data sets, which can be seen in our experiments involving binary classifications generated by arbitrary polynomials.  But this greater expressiveness comes at a cost:  it means that MOEs might be more prone to \textit{overfitting} the training data.  Overfitting is the phenomenon where a machine learning model fails to generalize from its training data distribution to samples outside of that distribution; a model that overfits is one that ``memorized'' the training data instead of learning the correct inductive rule.  This poses an inductive problem that we have not solved here.  Namely, the greater functional capacity of MOEs should be more prone to overfitting.  Whether they do or not depends on the data they are trained on, and the problem before them.  It is an open question, which we leave to future research, whether this is less of an issue because most real-world problems require the kind of higher functional capacity found in MOEs than in ensembles.  Maybe the world is just really complex---rendering the worry about overfitting an unfulfilled hypothetical.  On that issue, we remain silent here.
\section*{Appendix 1:  Proof of Proposition \ref{prop:moe}}\label{sec:appendix2}

Suppose $\mathcal{H}$ is a top-expert mixture of experts model of $n$ homogenous binary classifier experts drawn from the set $\mathcal{E}$ that is closed under translation with $VCD(\mathcal{E}) = m$ defined over instance space $\mathbb{R}$.  We need to show that there exists a set of points $X = \{x^{(i)} \in \mathbb{R}: x^{(1)}, \dots, x^{(nm)}\}$ such that $\mathcal{H}$ shatters $X$.  Recall that each $h \in \mathcal{H}$ is a piece-wise function defined as follows where $Z_{1}, Z_{2}, \dots, Z_{n}$ is a partition of $\mathbb{R}$:

\begin{equation*}
     h(x; n) = \begin{cases} 
      e_{1}(x) & x \in Z_{1} \\
      e_{2}(x) & x \in Z_{2} \\
      \vdots & \vdots \\
      e_{n}(x) & x \in Z_{n} 
   \end{cases}
\end{equation*}

\noindent where each $e_{i} \in \mathcal{E}$ is a binary classifier, $e_{i}: \mathbb{R} \to \{1,0\}$ for $i = 1, 2, \dots, n$.  Since $VCD(\mathcal{E}) = m$, there exists a set of points $X^{1}_{m} = \{x^{(i)} \in \mathbb{R}: x^{(1)}, \dots, x^{(m)}\}$ of cardinality $m$ that $\mathcal{E}$ shatters, i.e. for any labels $Y^{1}_{m} = \{y^{(1)}, \dots, y^{(m)}\}$, there exists $e \in \mathcal{E}$ such that $e(x^{(i)}) = y^{(i)}$ for $i = 1, \dots, m$.

What we need to show is that we can construct pairwise disjoint sets $X^{1}_{m}, \dots,$ $X^{n}_{m}$ such that for any labels $Y$ there exists some $h$ that correctly assigns the labels.  First, we build these pairwise disjoint sets.  Intuitively, the idea is that we use our initial set $X^{1}_{m}$ and shift it around.  We construct this set as follows:

\begin{equation*}
    X^{i}_{m} = \{x \in \mathbb{R}: \exists x^{(j)} \in X^{i-1}_{m} x = x^{(j)} + c_{i}\}
\end{equation*}

\noindent where each $c_{i}$ is chosen such that $x^{(j)} + c_{i} \notin \overset{i-1}{\underset{k=1}{\bigcup}} X^{k}_{m}$ for all $j = 1, \dots, m$ and $c_{1} = 0$.  We pick the constant that changes every member of the previous set to a new number.

We claim these sets are pairwise disjoint.  Let $X^{i}_{m}$ and $X^{k}_{m}$ such that $i \neq k$.  Suppose $i > k$.  Then by definition, we will have chosen $c_{i}$ such that each member $x \in X^{i}_{m}$ is not equal to $x^{(j)} + c_{k}$ for all $j = 1, \dots, m$.  Similarly, suppose $k > i$, then we will have chosen $c_{k}$ such that each member $x \in X^{k}_{m}$ is not equal to $x^{(j)} + c_{i}$ for all $j = 1, \dots, m$.

Next we define classifiers $e^{Y^{k}}_{j} \in \mathcal{E}$ for $j = 1, \dots, n$ with respect to the set of labels $Y^{k} = \{y^{(1)}_{k}, \dots, y^{(m)}_{k}\}$ as simply removing all the translations $c_{l}$ for $l = 1, \dots, j$ and applying the classifier $e^{Y^{k}}$ that correctly labels our points in the set $x^{(i)} \in X^{1}_{m}$, $e^{Y^{k}}(x^{(i)}) = y^{(i)}_{k}$ for $i = 1, \dots, m$.  We define for our new set of points $x^{*{(i)}} \in X^{j}_{m}$ the classifier $e^{Y^{k}}_{j}$:

\begin{equation*}
    e^{Y^{k}}_{j}(x^{*^{(i)}}) = e^{Y^{k}}(x^{*^{(i)}} - \overset{j}{\underset{l=1}{\sum}}c_{l})
\end{equation*}

\noindent That is each classifier $e^{Y^{k}}_{j}$ labels using $e^{Y^{k}}$ by removing the translations that built the members of $X^{j}_{m}$.  Note that $e^{Y^{k}}_{1} = e^{Y^{k}}$ since $c_{1} = 0$.  We know by assumption that $e^{Y^{k}}$ must exist for $Y^{k}$ for the set $X^{1}_{m}$.  It is clear then that for any arbitrary labels $Y^{k}$ on set $X^{j}_{m}$, the classifier $e^{Y^{k}}_{j}$ will correctly label the members of $X^{j}_{m}$ because those same labels applied to the original set $x^{(i)} \in X^{1}_{m}$ will be correctly classified by $e^{Y^{k}}$ and we simply note that $x^{(i)} = x^{*^{(i)}} - \overset{j}{\underset{l=1}{\sum}}c_{l}$.  Since this new classifier is nothing more than a translation of the older classier $e^{Y^{k}}$, it will be in the set $\mathcal{E}$.

We claim that for the set $\overset{n}{\underset{j=1}{\bigcup}}X^{j}_{m}$, for any labels $Y = \{y^{(1)}, \dots, y^{(nm)}\}$, there exists $h(x^{(i)}; n) = y^{(i)}$ for $i = 1, \dots, nm$.  Partition $Y$ into subsets $Y^{j}_{m}$ for $j = 1, \dots, n$ corresponding to each $X^{j}_{m}$ where each label in $Y^{j}_{m}$ applies to an element in $X^{j}_{m}$ for $j = 1, \dots, n$.  We define our mixture of experts hypothesis $h_{Y}$ as follows.:

\begin{equation*}
     h_{Y}(x; n) = \begin{cases} 
      e^{Y^{1}}_{1}(x) & x \in X^{1}_{m} \\
      e^{Y^{2}}_{2}(x) & x \in X^{2}_{m} \\
      \vdots & \vdots \\
      e^{Y^{n-1}}_{n-1}(x) & x \in X^{n-1}_{m} \\
      e^{Y^{n}}_{n}(x) & x \in C 
   \end{cases}
\end{equation*}

\noindent where $C = X^{n}_{m} \cup \mathbb{R} - \overset{n-1}{\underset{j=1}{\bigcup}} X^{j}_{m}$ is the catch-all for everything outside of our disjoint set.
Then since each $e^{Y^{j}}_{j}$ can correctly label each $y^{(i)} \in Y^{j}_{m}$, we can label the union of $Y^{j}_{m}$ corresponding to the union of $X^{j}_{m}$ for $j = 1, \dots, n$.  Thus $h_{Y}$ will correctly label $Y$ and so there exists a $h \in \mathcal{H}$ that shatters $\overset{n}{\underset{j=1}{\bigcup}}X^{j}_{m}$, which by the fundamental theorem of counting has a cardinality of $nm$.

\section*{Appendix 2:  Experiment Details}\label{sec:appendix3}

All code can be found at \url{https://github.com/brushing-git/PeirceMachine}.

Our linear regression experiments were conducted on datasets generated by univariate polynomials of degrees $[1,2,3,4,5]$. Feature data was drawn from a uniform distribution in the range $[-2.0,1.0]$, and we computed the targets via polynomials with coefficients drawn from the list $[2,3,-1,-1,1,1]$, e.g. the degree one polynomial was $p(x) = 2x + 3$, degree two was $p(x) = 2x^{2} + 3x - 1$, and so on. The data was then rescaled the data by subtracting the data mean $\mu$ and dividing by the standard deviation $\sigma$, $z(x) = (x - \mu) / \sigma$ to eliminate extreme values.  We then added noise from a normal distribution $\mathcal{N}(0, 0.1^{2})$ to the targets.  Finally, an identity feature was added.  The total size of the training data was $10000$ and the test data was $2000$ for each degree.

The models trained in the linear regression experiments include a Bayesian linear regression and MOE of two to three linear regressions.  The Bayesian linear regression assumed a conjugate prior, and we computed the posterior predictive distribution via \citep[399]{murphy2022probabilistic}:

\begin{align*}
    p(y | \mathbf{x}, \mathcal{D}) & = \int \mathcal{N}(y | \mathbf{x}^{\intercal}\theta, \sigma^{2}) \mathcal{N}(\theta | \mu, \Sigma) d\theta \\
    & = \mathcal{N}(y | \mathbf{x}^{\intercal}\mu, \mathbf{x}^{\intercal} \Sigma \mathbf{x} + \sigma^{2})
\end{align*}

\noindent where $\mu = \Sigma(\Sigma_{0}^{-1}\mu_{0} + \sigma^{-2}\mathbf{X}^{\intercal}\mathbf{Y})$ is the posterior mean and $\Sigma = \Sigma_{0}^{-1} + \sigma^{-2}\mathbf{X}^{\intercal}\mathbf{X}$ is the posterior covariance, and $\sigma$ is the known likelihood standard deviation.  Here $\mathbf{X}$ is the matrix of training data features $\mathcal{X}$, $\mathbf{Y}$ is the matrix of training data targets $\mathcal{Y}$, $\mu_{0}$ the prior mean, and $\Sigma_{0}$ the prior covariance.  Our prior was the standard normal $\mathcal{N}(0,I)$.  For the MOEs (more below), we trained sparse, top-2 expert MOEs using stochastic gradient descent and the Adam optimizer.  In training, the expert linear regressions learned to fit their means to the training data, and we used the likelihoods at inference time.  Our learning rate was set via an exponential decay schedule, where the learning rate for each epoch $\eta(x) = \eta_{0} \exp{(-\gamma x)}$, with a $\gamma = 0.75$ and an initial learning rate $\eta_{0} = 0.2$.  MOEs were trained for $30$ epochs.

The MOE models in all experiments were sparse, top-k MOEs \citep{shazeer2017outrageously}.  The key innovations in sparse, top-k MOEs are the use of only the top-k experts and the application of noise to the gating function.  Here our gating function $G$ is given by:

\begin{equation*}
    G(\mathbf{x}) = Softmax(KeepTopK(H(\mathbf{x}), k))
\end{equation*}

\noindent where $Softmax (\mathbf{x}) = \frac{\exp{(x_{i})}}{\sum^{k}_{j=1} \exp{(x_{j})}}$ and $KeepTopK$ and $H$ are as follows:

\begin{equation*}
    KeepTopK(v, k)_{i} = \begin{cases}
        v_{i} & \text{if $v_{i}$ is in the top $k$ elements of $v$} \\
        - \infty & \text{otherwise}
    \end{cases}
\end{equation*}

\begin{equation*}
    H(\mathbf{x})_{i} = (W_{g}\mathbf{x})_{i} + \epsilon \cdot \log (1 + \exp{((W_{noise}\mathbf{x})_{i}}))
\end{equation*}

\noindent where $\epsilon \sim \mathcal{N}(0,\mathbb{I})$, $W_{g}$ is a matrix of gate weights, and $W_{noise}$ is a matrix of noise weights.  The weights are learned during training.  For the VC dimension experiments, the MOEs were trained with the Adam optimizer and a fixed learning rate of $0.001$.

In the VC dimension experiments, we generated the data as described in section \ref{sec:experiments}.  Our feature data was constructed by sampling first $x_{1}$ from a uniform distribution in the interval $[-3.0, 3.0]$ and then computing $x_{2}$ by polynomials $p(x; m)$ of the correct degree $m \in [1,2,3,4,5,6,7,8]$ with coefficients sampled from $\mathcal{N}(0, 10)$ on $x_{1}$ and adding some noise to $x_{2}$ from $\mathcal{N}(0, \sigma)$, where $\sigma$ is the standard deviation of $x_{2}$.  We then rescaled the features identically to the linear experiments and added an identity feature.  The targets were computed from the non-identity features by thresholding the feature $x_{2}$ based on the computed polynomial $p(x; m)$.  The total training data size was $10000$ and the test set size was $2000$ for each degree of polynomial.

For the VC dimension experiments themselves, we trained two Bayesian logistic regressions via stochastic gradient Hamiltonian Monte Carlo (SGHMC) \citep{chen2014stochastic} and variational inference (VI) \citep{blei2017variational}.  SGHMC is a Hamiltonian Monte Carlo method that approximates the posterior by taking noisy samples from the posterior energy function $U(\theta) = - \sum^{n}_{i=1} \log p(y_{i} | x_{i}, \theta) - \log p(\theta)$ since $p(\theta | \mathcal{D}) \propto \exp{(-U(\theta) / T)}$ where $T$ is a temperature parameter.  Traditional Hamiltonian Monte Carlo approximates from the whole dataset, but SGHMC approximates from mini-batches of size $n^{\prime} \ll n$, which corresponds to the approximate posterior energy function $\overset{\sim}{U}(\theta) = \frac{n^{\prime}}{|n|} \sum^{n^{\prime}}_{i=1} \log p(x_{i} | \theta) - \log p(\theta)$.  We update our parameters $\theta_{k}$ at each epoch $k$ by adding a momentum term $v_{k-1}$ $\theta_{k} = \theta_{k-1} + v_{k-1}$, where $v_{k}$ is updated at each epoch by using the gradient of the approximate posterior energy function $v_{k} = v_{k-1} - \alpha_{k}\nabla\overset{\sim}{U}(\theta_{k}) - \eta v_{k-1} + \sqrt{2(\eta - \hat{\gamma})\alpha_{k}}\epsilon_{k}$ where $\alpha_{k}$ is the learning rate, $\eta$ is a friction parameter, $\hat{\gamma}$ is an estimate of the data noise, and $\epsilon \sim \mathcal{N}(0,\mathbb{I})$ is noise sampled from a normal distribution.  In our experiments, $\eta = 0.9$, the learning rate $\alpha_{k}$ was determined by an exponential decay schedule, $\hat{\gamma}$ was set to $10000^{-1}$, we trained the models for $84$ burn-in epochs (without any noise added), and we then collected $16$ samples for inference, which proceeds by an average over those samples.  VI is an approximation scheme where we substitute an intractable posterior for a more manageable probability distribution called our variational distribution $q$ with parameters $\phi$.  We then train our model by minimizing the KL-divergence between our true posterior and the variational distribution, which can be done by minimizing the evidence lower bound (ELBO) $\mathcal{L}(x, \phi) = \mathbb{E}_{z \sim q(z; \phi)}[\log p(x,z) - T\log q(z; \phi)]$, where $T$ is the temperature parameter.  One popular technique is the Bayes By Backprop \citep{blundell2015weight} method, where we use the reparameterization trick to compute the gradients for the ELBO via the backpropagation algorithm.  We use this method with a standard normal prior $\mathcal{N}(0,\mathbb{I})$ on model parameters and temperature of $0.1$.  Our models were trained for $100$ epochs at a fixed learning rate of $0.01$.  At inference time, we draw $16$ samples from VI logistic regressions to approximate the predictive posterior via an average.

\bibliography{main}
\bibliographystyle{apalike}

\end{document}